\documentclass[11pt]{article}

\usepackage{amsmath, mathtools, amsthm, graphicx, float, bm, csquotes}
\usepackage{csquotes}
\usepackage{setspace}
\usepackage{multirow}
\usepackage{booktabs}
\usepackage{makecell}
\usepackage[backend=biber,style=authoryear, sorting=nyt, citestyle=authoryear, doi=false,isbn=false,url=false]{biblatex}
\usepackage[colorlinks=true,linkcolor=blue,citecolor=blue]{hyperref}
\usepackage{times}
\usepackage{graphicx}
\usepackage{hyphenat}

\begin{document}

\title{raceBERT -- A Transformer-based Model for Predicting Race and Ethnicity from Names\footnote{https://github.com/parasurama/raceBERT}}
\author{Prasanna Parasurama\thanks{pparasurama@gmail.com} \\ New York University}
\date{\today}
\maketitle

\begin{abstract}
 This paper presents raceBERT -- a transformer-based model for predicting race and ethnicity from character sequences in names, and an accompanying python package.
 Using a transformer-based model trained on a U.S. Florida voter registration dataset\footnote{I sincerely thank Gaurav Sood and Suriyan Laohaprapanon for sharing their data}, the model predicts the likelihood of a name belonging to 5 U.S. census race categories (White, Black, Hispanic, Asian \& Pacific Islander, American Indian \& Alaskan Native).
 I build on \textcite{sood_predicting_2018} by replacing their LSTM model with transformer-based models (pre-trained BERT model, and a roBERTa model trained from scratch), and compare the results.
 To the best of my knowledge, raceBERT achieves state-of-the-art results in race prediction using names, with an average f1-score of 0.86 -- a 4.1\% improvement over the previous state-of-the-art, and improvements between 15-17\% for non-white names.
\end{abstract}

\section{Introduction}

Researchers studying racial disparities often do not have self-reported demographic data readily available, and must rely on proxies such as name and location to predict race (e.g. \textcite{zhang_assessing_2018,fiscella_use_2006}).
Similary, researchers studying racial discrimination are often interested in racial \textit{signal} encoded in names \parencite{bertrand_are_2004,kang_whitened_2016}. 
In hiring discrimination studies, for example, the recruiter's perception of a candidate's race (as proxied by name) becomes important, in which case it's useful to estimate the likelihood of name belonging to a particular race \parencite{parasurama_who_2020}. 
Over the years, new methods and models have been proposed to incrementally improve the accuracy of race prediction models \parencite{fiscella_use_2006,imai_improving_2016,ambekar_name-ethnicity_2009,sood_predicting_2018,xie_rethnicity_2021}.
This paper contributes to this line of research by presenting a transformer-based race and ethnicity prediction model, which, to the best of my knowledge, achieves state-of-the-art results in predictive accuracy. 

\section{Data}
The race prediction model uses the U.S. Florida voter registration dataset\footnote{https://dataverse.harvard.edu/dataset.xhtml?persistentId=doi:10.7910/DVN/UBIG3F} collected by \textcite{sood_predicting_2018}.
In total, there are 13,089,545 names across 5 race categories.
Table \ref{tab:race_counts} reports the counts and label for each race category.
Note that \texttt{unknown}, \texttt{other}, and \texttt{multiracial} categories are dropped, because there are too few examples.

\begin{table}[H]
 \centering
 \begin{tabular}{llr}
    \toprule
    Race                              & Label     & Count     \\
    \midrule
    Non-hispanic White                & nh\_white & 8,757,268 \\
    Hispanic                          & hispanic  & 2,179,106 \\
    Non-hispanic Black                & nh\_black & 1,853,690 \\
    Asian \& Pacific Islander         & api       & 253,808   \\
    American Indian \& Alaskan Native & aian      & 45,673    \\
    \bottomrule
\end{tabular}

 \caption{Florida voter registration data race label and counts}
 \label{tab:race_counts}
\end{table}

\section{Models}
The main point of departure from \textcite{sood_predicting_2018} is that I replace their LSTM classifier with transformer-based models.

\subsection{Fine-tuning a pre-trained BERT model}

As a baseline, I start with a pretrained BERT\footnote{https://huggingface.co/bert-base-uncased} model for sequence classification \parencite{vaswani_attention_2017,devlin_bert_2019}.
Although BERT classification typically operates at the sentence level with sequences of tokens, it can also be used at the token level with the sequences of word pieces (i.e. sequences of characters).
First, in the preprocessing step, the name is lowercased, and the first name and the last name are concatenated by an underscore.
For example \texttt{George Smith} becomes \texttt{george\_smith}.
Then the normalized name gets tokenized into words or word pieces depending on whether the name is in the model's vocabulary. 
For example, \texttt{george\_smith} gets tokenizen into \texttt{{[[CLS], george, \_, smith, [SEP]]}}.
Here, \texttt{george} and \texttt{smith} become distinct tokens because both names are part of the vocabulary. 
If a name is not in the vocabulary, it gets tokenized into word pieces. 
For example, \texttt{satoshi} and \texttt{nakamoto} are not in the vocabulary, therefore \texttt{satoshi\_nakamoto} gets tokenized into \texttt{[[CLS], sato, \#\#shi, \_, nak, \#\#amo, \#\#to, [SEP]]}

For the training step, following \textcite{sun_how_2019}, I use the following hyperparameters: \texttt{N\_EPOCHS=4, BATCH\_SIZE\_PER\_GPU=128, LEARNING\_RATE=2e-5, WEIGHT\_DECAY=2e-5}. 
See the \href{https://wandb.ai/parasu/raceBERT-public/runs/39jr1hrc/overview}{wandb project page} for a complete list of hyperparameters and configs. 

Table \ref{tab:pretrained_race_model_performance} reports the performance of the pre-trained model on a hold-out test set, and Table \ref{tab:pretrained_race_model_performance_comparison} compares raceBERT's performance to ethnicolr's performance (Note that ethnicolr does not have \texttt{aian} as a category). 
raceBERT achieves better performance across all race categories compared to ethnicolr.
On average, there is a 4.4\% improvement in the f1-score, with the highest performance improvement coming from non-white names, with improvements as much as 18\% for black names.

\begin{table}[H]
 \centering
 \begin{tabular}{lrrr}
    \toprule
    Group     & Precision & Recall & f1-score \\
    \midrule
    nh\_white & 0.89      & 0.94   & 0.91     \\
    nh\_black & 0.80      & 0.58   & 0.67     \\
    hispanic  & 0.85      & 0.88   & 0.87     \\
    api       & 0.79      & 0.66   & 0.72     \\
    aian      & 0.60      & 0.01   & 0.02     \\
    \midrule
    Average  & 0.87      & 0.87   & 0.86     \\
    \bottomrule
\end{tabular}

 \caption{raceBERT hold-out performence metrics (pre-trained language model)}
 \label{tab:pretrained_race_model_performance}
\end{table}

\begin{table}[H]
 \centering
 \begin{tabular}{lrrrrr}
    \toprule
    Group     & Precision & Recall & f1-score & \makecell[rb]{\texttt{ethnicolr}                 \\ f1-score} & \% Improvement \\
    \midrule
    nh\_white & 0.89      & 0.94   & 0.91     & 0.90               & 1.51           \\
    nh\_black & 0.80      & 0.58   & 0.67     & 0.55               & 18.15          \\
    hispanic  & 0.85      & 0.88   & 0.87     & 0.75               & 13.53          \\
    api       & 0.80      & 0.66   & 0.72     & 0.60               & 16.89          \\
    \midrule
    Average       & 0.87      & 0.87   & 0.87     & 0.83               & 4.38           \\
    \bottomrule
\end{tabular}

 \caption{raceBERT hold-out performence improvement (pre-trained language model)}
 \label{tab:pretrained_race_model_performance_comparison}
\end{table}

I repeat the training process with the Wikipedia ethnicity dataset \parencite{ambekar_name-ethnicity_2009,sood_predicting_2018} and report the results in Appendix A. 
As with the race model, raceBERT achieves an overall 5\% improvement over ethnicolr. 

\subsection{Training a model from scratch}

Although the pre-trained BERT model achieves significant performance improvements over ethnicolr, a few limitations remain.
First, with a vocabulary size of 30,000, the model is needlessly large for the task at hand. 
Most of the tokens in the vocabulary will never be used.
Second, many commonly occurring names are in the vocabulary, which raises the question of whether the model is simply memorizing names rather than learning from character sequences of names, in which case the generalizability of the model will suffer. 
To overcome these issues, I train a roBERTa model from scratch with a much smaller vocabulary size of 500 \parencite{liu_roberta_2019}. 

In the preprocessing step, the first name is lowercased, and the last name is uppercased, which are then concatenated by a space.
For example \texttt{George Smith} becomes \texttt{george SMITH}. 
In theory, the mixed lower/upper case will make it easier for the model to discriminate between first and last names.
Using the transformed names, I train a Byte Pair Encoding tokenizer with a max vocabulary size of 500 to learn the most commonly occurring character sequences, which is then used to tokenize all names.
For example, \texttt{george SMITH} gets tokenized into \texttt{[[CLS], ge, or, ge, SMITH, [SEP]]}.
Likewise, \texttt{satoshi NAKAMOTO} gets tokenized into \texttt{[[CLS], sa, t, o, sh, i, N, A, K, AM, O, T, O, [SEP]]}.

Next, I train a masked language model to learn the character sequences using the roBERTa masked language architecture \texttt{(ATTENTION\_HEADS=12, HIDDEN\_LAYERS=6)}. 
Using the weights from the language model, I initialize a roBERTa sequence classification model and train using the following hyperparameters: \texttt{N\_EPOCHS=4, BATCH\_SIZE\_PER\_GPU=128. LEARNING\_RATE=2e-5, WEIGHT\_DECAY=2e-5}.
All other configs can be found at the \href{https://wandb.ai/parasu/raceBERT-public/runs/39jr1hrc/overview}{wandb project page}.

Table \ref{race_model_performance} reports the performance of the model on a hold-out test set, and \autoref{race_model_performance_without_aian} reports the performance improvement compared to ethnicolr.
The model's performance metrics are almost the same as the pre-trained BERT model, but with a much smaller vocabulary and (theoretically) greater generalizability.
As such, I use this as the default model for the python package.

\begin{table}[H]
 \centering
 \begin{tabular}{lrrr}
    \toprule
    Group     & Precision & Recall & f1-score \\
    \midrule
    nh\_white & 0.89      & 0.94   & 0.91     \\
    nh\_black & 0.79      & 0.57   & 0.66     \\
    hispanic  & 0.85      & 0.88   & 0.86     \\
    api       & 0.78      & 0.64   & 0.70     \\
    aian      & 0.53      & 0.01   & 0.01     \\
    \midrule
    Average   & 0.86      & 0.87   & 0.86     \\
    \bottomrule
\end{tabular}

 \caption{raceBERT hold-out performence metrics}
 \label{race_model_performance}
\end{table}

\begin{table}[H]
 \centering
 \begin{tabular}{lrrrrr}
\toprule
metric &  precision &  recall &    f1 &  ethnicolr f1 &  \% improvement \\
group    &            &         &       &               &                \\
\midrule
nh\_white &       0.89 &    0.94 &  0.91 &          0.90 &           1.37 \\
nh\_black &       0.80 &    0.57 &  0.67 &          0.55 &          17.39 \\
hispanic &       0.85 &    0.88 &  0.87 &          0.75 &          13.32 \\
api      &       0.79 &    0.64 &  0.71 &          0.60 &          15.11 \\
\midrule
Average      &       0.87 &    0.87 &  0.87 &          0.83 &           4.11 \\
\bottomrule
\end{tabular}

 \caption{raceBERT hold-out performence improvements}
 \label{race_model_performance_without_aian}
\end{table}

\subsection{Code, Configs, and Resources}
All of the training code, as well as the raceBERT python package code, is on \href{https://github.com/parasurama/raceBERT}{Github}\footnote{https://github.com/parasurama/raceBERT}.
The trained models are uploaded to the \href{https://huggingface.co/pparasurama/raceBERT}{huggingface hub}\footnote{https://huggingface.co/pparasurama/raceBERT}.
All training configurations, hyperparameters, and learning curves can be found at the \href{https://wandb.ai/parasu/raceBERT-public}{wandb project page}\footnote{https://wandb.ai/parasu/raceBERT-public}.

\section{Conclusion \& Limitations}
This paper presents a new transformer-based model for predicting race from names and demonstrates the performance improvements over existing state-of-the-art models.
One limitation of this model is that it's trained on the Florida voter registration dataset, which is not necessarily representative of the U.S. population. 
As such, the accuracy scores may vary when used on datasets from U.S. regions with different demographic distributions. 
Another limitation -- one that's shared by all race prediction models -- is that the model is not 100\% accurate.
While this is admissable when studying racial disparities in the aggregate, or when estimating the likelihood of a name belonging to a particular race, it's not advisable to use this model in applications where it is important to know individual race.
 
\pagebreak

\printbibliography

\pagebreak

\section*{Appendix}
\appendix

\section{Wikipedia Ethnicity Dataset}

\begin{table}[H]
 \centering
 \begin{tabular}{lr}
    \toprule
    Ethnicity                             & Count  \\
    \midrule
    GreaterEuropean,British               & 43,579 \\
    GreaterEuropean,WestEuropean,French   & 14,076 \\
    GreaterEuropean,WestEuropean,Italian  & 13,633 \\
    GreaterEuropean,WestEuropean,Hispanic & 11,365 \\
    GreaterEuropean,Jewish                & 11,224 \\
    GreaterEuropean,EastEuropean          & 9,318  \\
    Asian,IndianSubContinent              & 9,063  \\
    Asian,GreaterEastAsian,Japanese       & 8,425  \\
    GreaterAfrican,Muslim                 & 6,854  \\
    Asian,GreaterEastAsian,EastAsian      & 6,514  \\
    GreaterEuropean,WestEuropean,Nordic   & 5,242  \\
    GreaterEuropean,WestEuropean,Germanic & 4,634  \\
    GreaterAfrican,Africans               & 4,348  \\
    \bottomrule
\end{tabular}

 \label{tab:ethnicity_counts}
 \caption{Wikipedia dataset ethnicity categories and counts}
\end{table}

\begin{table}[H]
 \centering
 \scalebox{0.8}{\begin{tabular}{lrrrrr}
    \toprule
    Group                                 & Precision     & Recall        & f1-score      & \makecell[rb]{\texttt{ethnicolr}                 \\ f1-score} & \%Improvement \\
    \midrule
    Asian,GreaterEastAsian,EastAsian      & 0.87          & 0.85          & 0.86          & 0.82                             & 4.65          \\
    Asian,GreaterEastAsian,Japanese       & 0.92          & 0.88          & 0.90          & 0.90                             & 0.00          \\
    Asian,IndianSubContinent              & 0.82          & 0.79          & 0.81          & 0.77                             & 4.94          \\
    GreaterAfrican,Africans               & 0.65          & 0.52          & 0.58          & 0.49                             & 15.52         \\
    GreaterAfrican,Muslim                 & 0.66          & 0.76          & 0.71          & 0.67                             & 5.63          \\
    GreaterEuropean,British               & 0.82          & 0.87          & 0.85          & 0.82                             & 3.53          \\
    GreaterEuropean,EastEuropean          & 0.81          & 0.78          & 0.80          & 0.76                             & 5.00          \\
    GreaterEuropean,Jewish                & 0.57          & 0.53          & 0.55          & 0.47                             & 14.55         \\
    GreaterEuropean,WestEuropean,French   & 0.71          & 0.68          & 0.69          & 0.64                             & 7.25          \\
    GreaterEuropean,WestEuropean,Germanic & 0.54          & 0.56          & 0.55          & 0.46                             & 16.36         \\
    GreaterEuropean,WestEuropean,Hispanic & 0.74          & 0.73          & 0.73          & 0.70                             & 4.11          \\
    GreaterEuropean,WestEuropean,Italian  & 0.77          & 0.79          & 0.78          & 0.75                             & 3.85          \\
    GreaterEuropean,WestEuropean,Nordic   & 0.82          & 0.74          & 0.78          & 0.68                             & 12.82         \\
    \midrule
    \textbf{Average}                      & \textbf{0.77} & \textbf{0.77} & \textbf{0.77} & \textbf{0.73}                    & \textbf{5.19} \\
    \bottomrule
\end{tabular}
}
 \label{tab:ethnicty_performance}
 \caption{raceBERT hold-out performance on Wikipedia ethnicity dataset}
\end{table}

\end{document}